\documentclass[letterpaper, 10 pt, conference]{ieeeconf}  
\usepackage{amsmath}
\usepackage{mathtools}
\usepackage{multirow}
\usepackage{bm}
\usepackage{xcolor}
\usepackage{subfigure}
\usepackage{url}   
\usepackage{algorithmic}
\usepackage{centernot}

\IEEEoverridecommandlockouts                              

\title{\LARGE \bf
Task-Specific Design Optimization and Fabrication for\\ 
Inflated-Beam Soft Robots with Growable Discrete Joints
}

\author{Ioannis Exarchos$^{1}$, Karen Wang$^{1}$, Brian H.~Do$^{2}$, Fabio Stroppa$^{3}$, Margaret M.~Coad$^{4}$,\\ Allison M.~Okamura$^{2}$, and C.~Karen Liu$^{1}$
\thanks{This work was supported in part by National Science Foundation grant 2024247, a National Science Foundation Graduate Research Fellowship, and an ARCS Foundation Fellowship.}
\thanks{$^{1}$I. Exarchos, K. Wang and C. K. Liu are with the Department of Computer Science, Stanford University, Stanford, CA 94305, USA. {\tt\small exarchos@stanford.edu, karenw24@stanford.edu, karenliu@cs.stanford.edu}}%
\thanks{$^{2}$B.~H.~Do, and A.~M.~Okamura are with the Department of Mechanical Engineering, Stanford University, Stanford, CA 94305, USA. {\tt\small brianhdo@stanford.edu, aokamura@stanford.edu}}%
\thanks{$^{3}$F.~Stroppa is with the Faculty of Computer Engineering, Kadir Has University, Turkey. {\tt\small fabio.stroppa@khas.edu.tr}}%
\thanks{$^{4}$M.~M.~Coad is with the Department of Aerospace and Mechanical Engineering, University of Notre Dame, Notre Dame, IN 46556, USA. {\tt\small mcoad@nd.edu}}%
}

\usepackage[ruled,vlined,commentsnumbered,titlenotnumbered]{algorithm2e}
\usepackage{mdframed}

\begin{document}

\definecolor{orange}{rgb}{0.9, 0.3, 0.0}
\definecolor{darkblue}{rgb}{0.0, 0.0, 0.53}
\definecolor{darkgreen}{rgb}{0.0, 0.53, 0.0}
\newcommand{\old}[1]{\textcolor{red}{\textbf {\st{#1}}}}
\newcommand{\new}[1]{\textcolor{red}{#1}}
\newcommand{\note}[1]{\cmt{Note: #1}}
\newcommand{\karen}[1]{\textcolor{red}{{[Karen: #1]}}}
\newcommand{\yannis}[1]{\textcolor{orange}{{[Keenon: #1]}}}
\newcommand{\js}[1]{\textcolor{blue}{{[JS: #1]}}}
\newcommand{\dalton}[1]{\textcolor{darkgreen}{{[Dalton: #1]}}}
\newcommand{\amy}[1]{\textcolor{purple}{[Amy: #1]}}
\newcommand{\eqnref}[1]{Equation~(\ref{eqn:#1})}
\long\def\ignorethis#1{}

\newcommand{\sect}[1]{Section~\ref{#1}}
\newcommand{\myparagraph}[1]{\vspace{5pt}\noindent\textbf{#1}}
\makeatletter
\DeclareRobustCommand\onedot{\futurelet\@let@token\@onedot}
\def\@onedot{\ifx\@let@token.\else.\null\fi\xspace}

\def\iid{i.i.d\onedot}
\def\eg{e.g\onedot} \def\Eg{E.g\onedot}
\def\ie{i.e\onedot} \def\Ie{I.e\onedot}
\def\cf{\emph{c.f}\onedot} \def\Cf{\emph{C.f}\onedot}
\def\etc{etc\onedot} \def\vs{vs\onedot}
\def\wrt{w.r.t\onedot} \def\dof{d.o.f\onedot}
\def\etal{et al\onedot}
\makeatother

\newcommand{\figtodo}[1]{\framebox[0.8\columnwidth]{\rule{0pt}{1in}#1}}
\newcommand{\figref}[1]{Figure~\ref{fig:#1}}
\newcommand{\secref}[1]{Section~\ref{sec:#1}}

\newcommand{\vc}[1]{\ensuremath{\boldsymbol{#1}}}
\newcommand{\pd}[2]{\ensuremath{\frac{\partial{#1}}{\partial{#2}}}}
\newcommand{\pdd}[3]{\ensuremath{\frac{\partial^2{#1}}{\partial{#2}\,\partial{#3}}}}

\newcommand{\vEndEff}{\ensuremath{\vc{d}}}
\newcommand{\vRelMove}{\ensuremath{\vc{r}}}
\newcommand{\sSet}{\ensuremath{S}}

\newcommand{\vControl}{\ensuremath{\vc{u}}}
\newcommand{\vPoint}{\ensuremath{\vc{p}}}
\newcommand{\sSpringCoef}{{\ensuremath{k_{s}}}}
\newcommand{\sDamperCoef}{{\ensuremath{k_{d}}}}
\newcommand{\vHandle}{\ensuremath{\vc{h}}}
\newcommand{\vForce}{\ensuremath{\vc{f}}}

\newcommand{\mTransChain}{\ensuremath{\vc{W}}}
\newcommand{\mRotateTrans}{\ensuremath{\vc{R}}}
\newcommand{\sJoint}{\ensuremath{q}}
\newcommand{\vJoint}{\ensuremath{\vc{q}}}
\newcommand{\mJoint}{\ensuremath{\vc{Q}}}
\newcommand{\mMass}{\ensuremath{\vc{M}}}
\newcommand{\sMass}{\ensuremath{{m}}}
\newcommand{\vGravity}{\ensuremath{\vc{g}}}
\newcommand{\vConstr}{\ensuremath{\vc{C}}}
\newcommand{\sConstr}{\ensuremath{C}}
\newcommand{\vCOM}{\ensuremath{\vc{x}}}
\newcommand{\sGeneralForce}[1]{\ensuremath{Q_{#1}}}
\newcommand{\vStateVar}{\ensuremath{\vc{y}}}
\newcommand{\vControlVar}{\ensuremath{\vc{u}}}
\newcommand{\argmax}{\operatornamewithlimits{argmax}}
\newcommand{\argmin}{\operatornamewithlimits{argmin}}

\newcommand{\tr}[1]{\ensuremath{\mathrm{tr}\left(#1\right)}}

\newcommand{\Ad}[1]{\text{Ad}_{#1}}
\newcommand{\AdTi}{\text{Ad}_{T_{\lambda(i),i}^{-1}}}

\newcommand{\dAd}[1]{\text{Ad}_{#1}^{*}}
\newcommand{\dAdTi}{\text{Ad}_{T_{\lambda(i),i}^{-1}}^{*}}

\newcommand{\ad}[1]{\text{ad}_{#1}}
\newcommand{\dad}[1]{\text{ad}_{#1}^{*}}

\newcommand{\I}{G}  
\newcommand{\AI}{\hat{G}} 
\newcommand{\AB}{\hat{B}} 

%
%

\renewcommand{\choose}[2]{\ensuremath{\left(\begin{array}{c} #1 \\ #2 \end{array} \right )}}

\newcommand{\gauss}[3]{\ensuremath{\mathcal{N}(#1 | #2 ; #3)}}

\newcommand{\pctab}{\hspace{0.2in}}
\newenvironment{pseudocode} {\begin{center} \begin{minipage}{\textwidth}
                             \normalsize \vspace{-2\baselineskip} \begin{tabbing}
                             \pctab \= \pctab \= \pctab \= \pctab \=
                             \pctab \= \pctab \= \pctab \= \pctab \= \\}
                            {\end{tabbing} \vspace{-2\baselineskip}
                             \end{minipage} \end{center}}
\newenvironment{items}      {\begin{list}{$\bullet$}
                              {\setlength{\partopsep}{\parskip}
                                \setlength{\parsep}{\parskip}
                                \setlength{\topsep}{0pt}
                                \setlength{\itemsep}{0pt}
                                \settowidth{\labelwidth}{$\bullet$}
                                \setlength{\labelsep}{1ex}
                                \setlength{\leftmargin}{\labelwidth}
                                \addtolength{\leftmargin}{\labelsep}
                                }
                              }
                            {\end{list}}
\newcommand{\newfun}[3]{\noindent\vspace{0pt}\fbox{\begin{minipage}{3.3truein}\vspace{#1}~ {#3}~\vspace{12pt}\end{minipage}}\vspace{#2}}

\newcommand{\norm}[1]{\left\lVert#1\right\rVert}

\newcommand{\key}{\textbf}
\newcommand{\fun}{\textsc}



\maketitle
\thispagestyle{empty}
\pagestyle{empty}

\begin{abstract}

Soft robot serial chain manipulators with the capability for growth, stiffness control, and discrete joints have the potential to approach the dexterity of traditional robot arms, while improving safety, lowering cost, and providing an increased workspace, with potential application in home environments. This paper presents an approach for design optimization of such robots to reach specified targets while minimizing the number of discrete joints and thus construction and actuation costs. We define a maximum number of allowable joints, as well as hardware constraints imposed by the materials and actuation available for soft growing robots, and we formulate and solve an optimization problem to output a planar robot design, i.e., the total number of potential joints and their locations along the robot body, which reaches all the desired targets, avoids known obstacles, and maximizes the workspace. We demonstrate a process to rapidly construct the resulting soft growing robot design. Finally, we use our algorithm to evaluate the ability of this design to reach new targets and demonstrate the algorithm's utility as a design tool to explore robot capabilities given various constraints and objectives.

\end{abstract}

\section{Motivation}

Soft robots are often created with the goal of offering compliant physical interactions between a robot and its environment, which can be safer and more readily accommodate uncertainty compared to traditional rigid robots. Moreover, the materials and actuation technologies used for soft robots can be less expensive and easier to manufacture or assemble than those used for traditional rigid robots. Soft robots have the potential to be designed with both mechanical properties and geometries appropriate for various tasks and environments. However, soft robots are typically limited in their dexterity because continuum arms lack the localized, large angle changes of traditional joints.

An attractive vision for soft robots is that regular people (i.e., not professional robot designers) can effectively create new robots to help them with tasks. This ``robot for every task,'' approach is motivated by the goal of democratizing the technology and its benefits across a wide spectrum of users \cite{SchulzRobogami2017}, as well as developing robotic devices that are uniquely suited to the needs of particular users~\cite{MorimotoABME2018}. Design optimization can automate the process of new robot creation. 

\begin{figure}[t!]
  \centering
	{\includegraphics[width=0.9\linewidth]{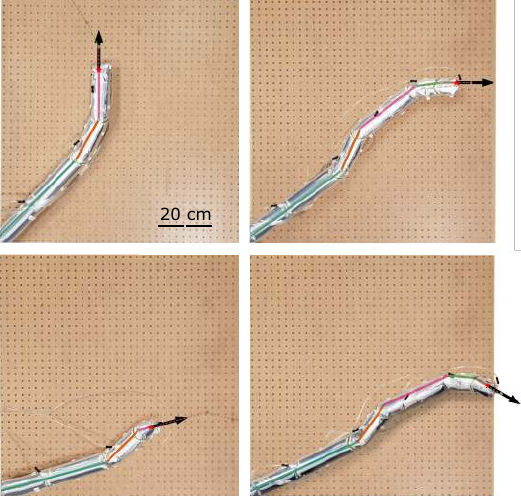}}
    \caption{A soft growing vine robot design with four discrete joints, optimized to achieve four different targets, each at a different orientation.Each colored segment indicates a link, and the joints are between the links. Due to growth, only the necessary joints are exposed for each target. Here, a \textit{single} robot design reaches all targets with their specified orientations.}
    \label{fig:intro}
    \vspace{-10pt}
\end{figure}

Toward the goal of creating soft robot manipulators that approach dexterity similar to that of traditional rigid robot arms while bringing advantages in compliance, low cost, and increased workspace, here we develop a new design optimization process coupled with the unique properties of soft growing ``vine'' robots~\cite{HawkesScienceRobotics2017} with stiffening capabilities and discrete joints~\cite{DoICRA2020} in order to create custom soft robot arms that can reach a set of specific targets, avoid known obstacles, maximize the workspace, while minimizing construction and actuation cost. The outcome of the design optimization is the specification for a soft serial-chain manipulator, which is then rapidly constructed and demonstrated to achieve the design goals, as shown in Fig.~\ref{fig:intro}.



\subsection{Related Work}

\subsubsection{Soft, inflated-beam, growing vine robots} Continuum robots characterized by tip extension, significant length change, and directional control are termed ``vine robots," due to their similar behavior to plants with the growth habit of trailing. In our instantiation, vine robots are inflated beams that extend in length from the tip using internal air pressure to pass the material of a flexible, tubular body through its center and turn it inside out at the tip (a process called eversion) \cite{HawkesScienceRobotics2017, BlumenscheinLM2017}. These robots can be made of low-cost materials, such as low-density polyethylene (LDPE) plastic, and can grow to very long lengths from a compact base, giving them a large workspace. They can also use active, reversible steering mechanisms based on cable tendons \cite{BlumenscheinRAL2018, BlumenscheinROBOSOFT2018, GanRAL2020} or pneumatic artificial muscles \cite{BlumenscheinROBOSOFT2018, GreerICRA2017, GreerSoRo2019, HawkesICRA2016, Naclerio2020ral} that are compatible with the growing mechanism. A number of researchers have recognized the advantages of tip growth for movement and manipulation in cluttered and constrained environments, because the robot body does not slide with respect to the environment~\cite{Haggerty2019IROS,Naclerio2018Burrow,OzkanAydin2019,sadeghi2017toward}.

In prior work, vine robot tip orientation has not been independent from its position~\cite{CoadRAM2020}. Here we build upon recent results~\cite{DoICRA2020, Wang2020icra, Selvaggio2020icra} that lock the robot's shape or use the environment to allow the robot tip to reach targets at different orientations. In particular, we use layer jamming to control the stiffness of vine robot sections~\cite{DoICRA2020} in order to create discrete joints at selected locations along the length of the robot. It is also possible to stiffen regions of the robot to improve shape control and payload handling, although this is beyond the scope this work~\cite{DoICRA2020}.

\subsubsection{Design optimization of soft robots}


Due to their infinite degrees of freedom and large design space that is linked to their physical properties, soft robots are prime candidates for design optimization.
Design can be interpreted as the set of parameters and characteristics used to build a soft robot, which might include lengths, actuators, joints, material properties, and local shape. Many methods for design optimization exist, including greedy algorithms \cite{KoehlerTRO2020}, sensitivity analysis \cite{megaro2017computational}, finite element analysis \cite{skouras2013computational}, supervised learning for sensor placement \cite{spielberg2021co}, geometric modeling to achieve desired deformations \cite{ma2017computational}, genetic algorithms for dexterity \cite{bodily2017multi}, gait parameters \cite{tesch2013expensive}, and counterweight balance \cite{coello1998using}. Furthermore, optimization has been used for control of soft robots, for example using reinforcement learning \cite{ansari2017multiobjective,coevoet2017optimization}.

Although design planning has been performed for passive, pre-shaped, everting tubes that exploit environmental contacts~\cite{GreerIJRR2020}, design optimization of actively steered vine robots has not yet been approached. In this work, we propose methods to balance competing demands on a robot design, including the limitations of actuation, while also considering geometric constraints that define the proposed task.



\section{Problem Statement}


\begin{figure}
\vspace{1.5mm}
    \centering
    \includegraphics[width=0.9\linewidth]{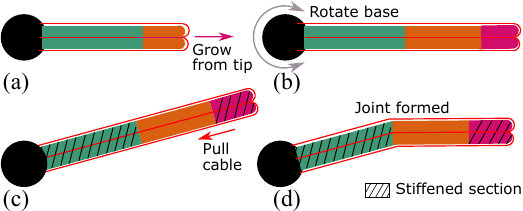}
    \caption{The vine robot has three distinct actuation modes. (a) Through pressure-driven eversion, the robot can add material to its tip. The robot can also shorten in length by retracting from the tip. (b) The material for this growth comes from a spool in a fixed based, which is free to rotate. (c) By stiffening certain sections while leaving others soft, a joint can be formed by pulling on a cable. (d) The angle of the joint is determined by the cable lengths and can be subsequently preserved by stiffening that section. Repeating (c) and (d) allows other joints to bend.}
    \label{fig:vineSchematic}
\vspace{-15pt}
\end{figure}

Fig. \ref{fig:vineSchematic} shows a schematic of the vine robot considered in this work. The robot originates from a fixed base from which it can grow from its tip or retract. The base itself is free to rotate. 
The robot body contains pre-fabricated sections whose stiffness can be independently controlled using layer jamming. By stiffening all sections except one and pulling on one of two cables in the planar case, the robot body bends at the proximal end of the softened segment, creating an effective revolute joint. When that section is then stiffened, the bend is held in place, then a different section is softened, and a cable is pulled to create a new bend. In general, we produce $n'$ discrete joints by fabricating $n'$ robot sections. Because of the ability to grow and retract, the vine robot can grow a different number of joints depending on how many have been everted, unlike conventional robots that have a constant number of joints. Those joints that have not been everted do not contribute to the robot shape.



Rotation by cable pulling introduces some mechanical limitations. Joint rotations near the tip of the robot are challenging due to small applied moments. Additionally, there may be a maximum bending angle less than self-intersection. Bending an inflated beam results in spring-like behavior where the beam produces a resistance torque proportional to the angle it is bent \cite{Nesler2018interference}. Depending on the maximum beam wall stiffness, the maximum torque the jammed layers can resist may depend on the angle of the joint. For our demonstrations, we choose [$-30^\circ$, $30^\circ$] as an estimate for the range of controllable angles.

A greater number of joints results in greater dexterity and a larger workspace. However, there is a trade-off between these attributes and the time and complexity of robot fabrication, as well as the complexity of controlling the robot once fabricated. For many applications with a predictable set of tasks, the locations of targets may be known ahead of time. Thus, our primary goal is to minimize the fabrication time and cost by producing a robot with the fewest number of joints while still reaching all targets. However, a fixed design is capable of reaching many more targets than those it was explicitly designed for, and situations may arise where we wish to reach targets whose locations are not known ahead of time. Thus, for a given fixed design, we also evaluate whether additional target positions/orientations can be achieved.


\section{Design Optimization}

Given the base location of the vine robot, a set of target locations and orientations for the end-effector, and a set of known obstacles, we wish to obtain an optimal design of the vine robot that reaches all the targets, avoids all fixed obstacles, covers the largest target space, and satisfies all hardware constraints. The parameters of design include the total number of the potential joints and their locations along the vine robot, which also determine the length of each link of the robot. These design choices need to be determined before the robot is manufactured and assembled. 


Unlike rigid robot arms for which the design space can be simply represented by the length of each link, the vine robot's design space needs to consider a set of unique characteristics owing to its ability to grow and retract: (1) Different targets might require different numbers of links to reach. (2) Any interior link has a fixed length across different targets while the length of the terminal link can vary for different targets, though not exceeding its maximum length (which is equal to the one it possesses as an interior link). (3) A link can be interior or terminal for different targets. (4) The number of links required for a given set of targets is unknown.

Figure \ref{fig:intro} illustrates one possible design that can reach four targets with the desired orientations. This design requires four potential joints to be placed at 0.44, 0.61, 0.88 and 0.98 meters from the base of the vine robot. For each target, the vine robot needs to expose a different number of joints to reach. The length of each link is fixed if the link is not the terminal one (the most distal one). For the terminal link, it can grow or shrink arbitrarily within the bounds. 

\subsection{Formulating the Optimization}
Given a budget of $n$ links of a vine robot, we wish to determine their lengths $\boldsymbol{\ell} \in \vc{R}_+^n$ and a set of bending angles $\vc{Q}=[\vc{q}_1^T, \vc{q}_2^T,\ldots,\vc{q}_m^T]$, such that $m$ pairs of target location and orientation, $\vc{T} = [(\vc{t}_1, \phi_1), \ldots, (\vc{t}_m, \phi_m)]$ can be reached with the same design while avoiding fixed obstacles $\vc{o} \in \mathcal{O}$ and maximizing the workspace. A design is defined solely by $\boldsymbol{\ell}$. A particular configuration $\vc{q}$ to reach a given target is not part of the design. The variables of optimization are $\boldsymbol{\ell}$ (shared across all targets) and $\vc{Q}=[\vc{q}_1^T, \vc{q}_2^T,\ldots,\vc{q}_m^T]$, the collection of all configurations, different for each target.

At first glance, this problem involves optimizing continuous variables (i.e., link lengths and joint angles) and solving for a discrete variable---how many joints do we need for each target? We can bypass answering this discrete question, thereby avoiding solving a challenging mixed-integer program, if we exploit the unique aspects of soft growing robots. Instead of following the conventional rigid robot design, in which the end-effector is always at the tip of the robot arm, we allow any point along the vine robot to be the end-effector. If a point $\vc{p}$ on the vine robot can reach the target with the desired orientation, we consider it a successful design for this target and discard all the materials and joints beyond $\vc{p}$. The fact that the vine robot has the ability to retract makes our formulation possible. Concretely, we solve for a vector $\boldsymbol{\ell} \in \vc{R}_+^n$, but the optimizer might end up only using $n' < n$ links to reach all $m$ targets and yielding a design with only $n'$ links. Our method results in a generic continuous optimization which automatically determines the optimal number of links required for each target. 

\subsubsection{Constraints} The hardware constraints are specified by a maximum bending angle at any given joint in the robot: with the exception of the root joint (i.e., the base of the robot) which is allowed to rotate freely, bending angles of all other joints are restricted to $[-30^\circ, 30^\circ]$. Another constraint is that the minimum possible length of a link is 10~cm, while we do not consider links longer than 1~m.  

\subsubsection{Objective function}
The geometric interpretation of a feasible solution requires targets to lie on the line segment corresponding to any one of those links, with the link orientation matching the prescribed target orientation and no collisions between any link and obstacle. In addition, we would like the design to maximize the workspace. 

\paragraph{Target location} The (minimum) distance of the target to the line segment represented by the link. The minimum distance from a target $\vc{t}_j$ to link $i$ of the current estimate of design is 
\[d(\bm{\ell}, \vc{q}_j, \vc{t}_j, i) = \|\vc{t}_j-\vc{t}^*(\bm{\ell}, \vc{q}_j,i)\|,\]
where $\vc{t}^*$ is either $\vc{t}_j$'s projection on link $i$, or one of link $i$'s endpoints, if the projection lies outside the link. Note that to discourage solutions with targets lying very close to a link node (which would lead to very small end-effector links), we prefer the target to lie somewhere between $30-90\%$ of the link length.

\paragraph{Target orientation} The absolute difference between the link orientation and the prescribed target orientation $\phi_j$, namely, \[o(\vc{q}_j, \phi_j, i) = |\phi_j - \sum_{k=1}^i \vc{q}_j(k)|,\] where $i$ is the link index we are considering and $\vc{q}_j$ is the current estimate of robot configuration for target $j$. 

These two quantities are weighted by $\beta$ to balance the difference in units. The weighted sum represents the cost for each individual link $i$, for a given target $(\vc{t}_j, \phi_j)$,
\begin{equation}
    L_{tar}(\bm{\ell}, \vc{Q}) = \sum_{j=1}^m \frac{\min_i \big(d(\bm{\ell}, \vc{q}_j, \vc{t}_j, i) \\ + \beta\; o(\vc{q}_j, \phi_j, i)\big)}{\|\vc{t}_j\|^2}.
\end{equation}

Taking the minimum across links yields the cost of the ``active end-effector'' link, i.e., the particular link that will be considered towards reaching the target. These procedures are repeated across all targets, thus producing a vector of costs. Those costs are then weighted by the inverse of the target squared distance from the origin (thus giving priority to closer targets which are more difficult to satisfy; this is because the bending restrictions force us to use fewer links to reach close targets, thus fewer degrees of freedom) and summed up to furnish the total cost. Note that the ``active end-effector'' link may change during the process of optimization.

\paragraph{Obstacle Violation} When attempting to reach a desired target, a cost will incur if any part of the robot intersects with an obstacle. We implement a collision checking function $b(\bm{\ell}, \bm{Q}, i, j, k)$ against two types of obstacles, polygons and circles. $b$ is a binary function that returns $1$ if link $i$ at pose $j$ is intersecting with obstacle $k$, and returns $0$ otherwise. The loss function for the obstacle violation can be defined as:
\begin{equation}
    L_{obs}(\bm{\ell}, \vc{Q}) =
    \begin{cases}
      c, & \text{if}\;\exists \;i,j,k \;\; \text{s.t.}\; b(\bm{\ell},\bm{Q},i,j,k) = 1 \\
      0, & \text{otherwise},
    \end{cases}
  \end{equation}
where $c$ is a large constant, indicating the penalty of obstacle violation. We set $c= 20$ in our implementation. 
\paragraph{Coverage of workspace} The design should also maximize the coverage space. Since an analytical solution for coverage is difficult to obtain, we opt for a Monte Carlo approximation, where we sample uniformly in the 3D target space of $(t_x, t_y, \phi)$ and check whether each sampled target is reachable. We define the coverage of a design, $g(\bm{\ell})$, as the ratio of reachable targets to the total sampled targets. Since we need to check reachability for a large number of targets every time we evaluate $g(\bm{\ell})$, we need a very fast routine to check whether a design can reach a sampled target. 

Algorithm~\ref{alg:Reach} checks the reachability using the first $n'$ links, where $n'$ iterates from $2$ to $|\bm{\ell}|$, the total number of links in the design. For each iteration, we determine a target segment $\overline{\vc{t} \vc{p}}$, where $\vc{p}$ is a point on the line extended backward from $\vc{t}$ aligned with $\phi$, subject to $\|\vc{t}-\vc{p}\| = \sum_{i=n'}^{|\bm{\ell}|} \ell_i$. We first check whether the most straight configuration ($\bm{q} = \bm{0}$) using the first $n'-1$ can intersect $\overline{\bm{t}\bm{p}}$. If so, adding $n'$-th link from the intersection point to $\bm{t}$ will allow the robot to reach the target at the desired orientation $\phi$. We still will have to check if the intersection angle is within the joint limit. If not, we bend the first $n'-1$ links of the robot to the joint limits ($\bm{q} = \bm{30}^{\circ} or -\bm{30}^{\circ}$) and check intersection angle with $\overline{\bm{t}\bm{p}}$. If the most bent configuration no longer intersects $\overline{\bm{t}\bm{p}}$, we unbend the links one by one starting with the first link. If at any point, the intersection angle is within the acceptable bending range, Alg.~\ref{alg:Reach} returns $\texttt{true}$ and exits. 
\begin{algorithm}[b!]
\caption{Reachable Check}
\KwIn{target $(\bm{t}, \phi)$ and link lengths $\bm{\ell}$}
\If {$|t| \leq \ell_0$} {
    \Return {$|\phi| < \epsilon$}
    }
\For {$n' = 2: |\bm{\ell}|$} {
    $\bm{p} = (t_x - \sum_{i = n'}^{|\bm{\ell}|} \ell_i \cdot \cos \phi, t_y - \sum_{i = n}^{|\bm{\ell}|} \ell_i \cdot \sin \phi)$ \\
    $\bm{q} = \bm{0}$ \\
    $(b, \theta)$ = CheckIntersect($\bm{t}, \bm{p}, \bm{\ell}, \bm{q}, n')$ \\
    \If {$b ==$ false} {
        continue}
    \If {$-30^\circ < \theta < 30^ \circ$} {  
        \Return {true}}
    \If {$\theta > 30^\circ$} {
        $\bm{q} = \bm{30^\circ}$ }
    \Else {
        $\bm{q} = \bm{-30^\circ}$ }
   \For {$j = 1:n'-1$} {
        $(b, \theta)$ = CheckIntersect$(\bm{t}, \bm{p}, \bm{\ell}, \bm{q}, n')$ \\
        \If {$b ==$ true and $|\theta| < 30^\circ$}  {
            \Return {true}
        }
        $q_j = 0$
        }
}
\Return {false}
\label{alg:Reach}
\end{algorithm} 

\begin{algorithm}[h!]
\caption{CheckIntersect}
\KwIn{ target $(\bm{t}, \phi)$, point $\bm{p}$, link lengths $\bm{\ell}$, link configurations $\bm{q}$, and number of links used $n'$}
\KwOut{existence of intersection $b$ and angle of intersection $\theta$}
$\bm{r} = (\sum_{i = 0}^{|n'-1|} \ell_i \cdot \cos \phi, \sum_{i = 0}^{|n'-1|} \ell_i \cdot \sin \phi)$\\
$\mathcal{S} = \overline{\bm{tp}} \cap \text{circle}(\bm{0},|\bm{r}|)$ \\
\If {$\mathcal{S} = \emptyset$} {
    \Return{ (false, 0)}}
$\bm{s}_0$ = closest$(\bm{t}, \mathcal{S})$ \\
$\theta = \arccos(\bm{s}_0, \bm{t-p})$ \\
\Return{ (\text{true}, $\theta$)}
\end{algorithm} 

 The cost function can be summarized by:
\begin{multline}
\label{eqn:costs}
L(\bm{\ell}, \vc{Q}) = w_t\;L_{tar}(\bm{\ell}, \vc{Q}) + w_b\; L_{obs}(\bm{\ell}, \vc{Q}) - w_g\; g(\bm{\ell}).
\end{multline}


\begin{algorithm}[h!]
\caption{Adaptive Stochastic Search for Design}
\textbf{Input:} objective function $L(\vc{x})$, where $\vc{x} =[\bm{\ell},\vc{q}_1, \vc{q}_2,\ldots,\vc{q}_m]$, initial guess $\bm{\mu}$ and $\bm{\sigma}$, max link budget $n_{max}$, learning rate $\alpha$, shape function $S$, number of samples $K$, number of iterations $N$, sampling variance lower bound $\varepsilon=10^{-3}$ \\
\For{$n=2, \ldots, n_{max}$ or feasible design found} {
Initialize $\bm{\ell}$ to be size $n$ \\ 
  \For{$i=1,\ldots,N$ or convergence criterion} {
    \For{$k=1,\ldots,K$} {
        $\vc{x}^k=\bm{\mu}_i + \Delta \vc{x}^k$, $\;\; \Delta \vc{x}^k \sim \mathcal{N}(0,\bm{\sigma_i}^2)$  \tcc{$\bm{\sigma}^2$ means element-wise multiplication} 
        Clip $\vc{x}^k$ for constraints \\
        $L^k=-L(\vc{x}^k)$; 
    }
    $L_{min} = \min_k L^k$,
    $L_{max} = \max_k L^k$ \\
    \For{$k=1,\ldots,K$} {
        $L^k = \frac{L^k -L_{min}}{L_{max} - L_{min}}$; \\
        $S^k=S(L^k)$
    }
    $\bm{\mu}_{i+1}=\bm{\mu}_i+\alpha \frac{\sum^{K}_{k=1}S^k\Delta \vc{x}^k}{\sum_{k=1}^K S^k}$ \\
    $\bm{\sigma}_{i+1} = \sqrt{(\sum_{k=1}^{K} S^k(\Delta \vc{x}^k)^2+\bm{\varepsilon)}}$ 
    Clip $\bm{\mu}_{i+1}$ for constraints }
  $\vc{x}^*=\bm{\mu}_{N}$
  }
\Return{$\vc{x}^*$}
\label{alg:SS}
\end{algorithm}

\subsection{Solving the Optimization}
The nature of the cost is highly nonlinear, non-convex, non-differentiable, and with a plethora of local minima. For these reasons, we opted for gradient-free, sampling-based optimization, specifically Adaptive Stochastic Search \cite{zhou2014stoch_search}. Adaptive stochastic search is a sampling-based method within stochastic optimization that transforms the original optimization problem via a probabilistic approximation. The core concept behind this algorithm is approximating the gradient of the objective function by evaluating random perturbations around some nominal value of the variable of optimization, a concept that also appears under the name Stochastic Variational Optimization and shares many similarities with natural evolution strategies and the Cross Entropy Method \cite{rubinstein2001combinatorial,de2005CEM}.  Given the objective function $L(\cdot)$ and an initial guess solution $\vc{x}$, where $\vc{x}=[\bm{\ell},\vc{q}_1, \vc{q}_2,\ldots,\vc{q}_m]$, we sample $K$ solutions using mean $\bm{\mu}=\vc{x}$ and variance $\bm{\sigma}^2$ (we abuse the notation here to indicate $\bm{\sigma}$ multiplying by itself element-wise), evaluate the objective function for each one of them. and map its value using a shape function $S(\cdot)$; in our implementation we used $S(L) = \exp({10L})$. We then normalize the values $S(L)$ in $[0,1]$ and update the mean and variance of our sampling through an $S$-weighted combination of the sampled solutions; thus, the better a sampled solution is, the more it influences the newly calculated mean and variance. To satisfy the constraints, samples and updated $\bm{\mu}$'s are clipped to the allowable range of values. The procedure is summarized in Alg.~\ref{alg:SS}. While \cite{zhou2014stoch_search} was developed to accommodate sampling from any distribution of the exponential family, we opted for Gaussian sampling for its simplicity. Furthermore, we applied clipping to ensure constraints are satisfied, and performed normalization of $L$ to improve convergence of the algorithm. However, there are no mathematical guarantees of convergence, as addressed in \cite{zhou2014stoch_search}. For our problem that has a plethora of local minima, while the sampling nature of Alg.~\ref{alg:SS} helps overcome some of these local minima, there is not theoretic guarantee for global optimality. 

The link budget $n$ is also a discrete variable, but we can avoid formulating a mixed-integer program using a simple 1-D search. That is, we optimize over $n$ by repeating the aforementioned optimization procedure starting from $n=2$ and increasing the budget until a solution that satisfies all targets/orientations is obtained (see the outer loop in Alg.~\ref{alg:SS}). This is practical because, due to the low dimensionality of the optimization problem for each $n$, a solution can be obtained relatively quickly. The code is available at {\small \url{https://github.com/iexarchos/SoftRobotDesOpt.git}}



\section{Evaluation}\label{Eval}

\subsection{Fabrication of an Optimal Design} 
We demonstrated the entire process of prototyping a vine robot from design optimization to hardware fabrication. We considered a base at $(0.0,0.0)$ and four targets located at $(0.4,0.65)$,  $(0.8,0.6)$, $(0.9,0.4)$, and $(0.6,0.25)$ (all units in meters), with corresponding prescribed end-effector orientations of $90^{\circ}$, $0^{\circ}$, $-30^{\circ}$, and $15^{\circ}$, respectively. The result of this optimization is shown later in Fig.~\ref{fig:des_comp}(a), in comparison to designs with different constraints.

The vine robot consists of a main body containing embedded pouches and a base through which the robot is actuated. Two cable tendons, on the left and right sides, are routed along the vine robot length and connected to motorized spools. The eversion and retraction of the robot are controlled using a motorized spool in the base. The base is connected to an air supply, which provides the pressure for growth. The robot body is fabricated from two LDPE tubes. Chevron-shaped pouches in the body are formed by heat sealing the tubes together. Layer stacks were then placed into the pouches and secured to the vine robot body using double-sided tape. A 1/8~inch outer diameter plastic tube was routed to each pouch, allowing for control of each pouch pressure. By connecting the pouch to the vine robot pressure source, pouch pressure equaled the vine robot body pressure and the layers were unjammed. By disconnecting the tube from the body pressure source, the layers were jammed.

\begin{figure}[t]
\vspace{1.5mm}
    \centering
    \includegraphics[width=0.9\linewidth]{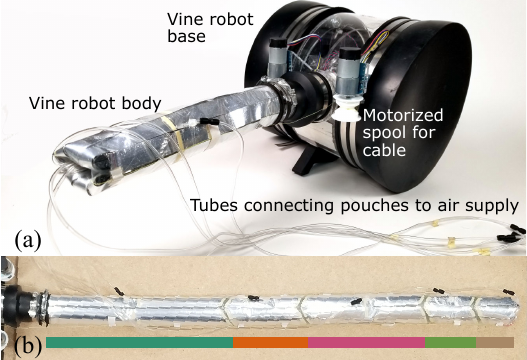}
    \caption{(a) The vine robot is actuated by two cables running along its length and controlled via two motorized spools. Pouch pressure is controlled via tubes connected to each pouch. (b) Vine robot segment lengths were fabricated using the optimized $\bm{\ell}$. Each color corresponds to a link $\ell_i$.}
    \label{fig:VineIsometric}
\vspace{-10pt}
\end{figure}

For a specified link budget $n$, the optimization returns a sequence of $n'$ link lengths $\bm{\ell}$ to construct. These link lengths correspond directly to the lengths of the layer stacks and pouches to be fabricated. Layers are cut into parallelograms and assembled into stacks of 14 layers by taping the ends of the layers together. We laser cut the layers from 0.05~mm thick aluminum-sputtered polyester film. 
To facilitate eversion and retraction, each pouch contains eight stacks of layers arranged circumferentially. The optimization in this example yielded a design consisting of five links, with link lengths $0.44$, $0.17$, $0.27$, $0.10$, and $0.11$ meters, ordered from the base to the tip. Fig.~\ref{fig:VineIsometric}(b) shows the fabricated vine robot. Fig.~\ref{fig:intro} qualitatively shows how the fabricated robot achieves four different targets in a plane; quantitative performance analysis will be performed in future work.

\subsection{Target Reachability Analysis}
To assess the algorithm performance in optimizing robot design and configurations, we tested it on randomly sampled targets and orientations. Specifically, we varied both the number of targets, $m$, and the link budget, $n$. This introduced the following difficulty: when picking a set of targets and orientations randomly, it is unclear whether a failure of the algorithm to find a solution given a link budget is due to the algorithm's performance, or because a feasible solution satisfying the constraints does not exist for the given link budget. To avoid this problem, we sampled $m$ targets and orientations by first sampling a random five-link solution with $m$ different configurations, and then obtaining the targets and configurations by selecting points along any of the links of the robot, along with their respective link orientation. In this way a feasible solution is guaranteed to exist as long as the link budget during optimization is $n\geq 5$. The results of this analysis are shown in Table~\ref{tab:res}. For the link budget above 4, our algorithm has a 100\% success rate in obtaining a solution for a feasible target.

\begin{table}[t]
\vspace{2mm}
\centering
\caption{Design optimization success rate}
\begin{tabular}{|c||c|c|c|c|c|}
\hline
\multirow{2}{*}{Link budget ($n$)} & \multicolumn{5}{c|}{\# targets ($m$)}                                     \\ \cline{2-6} 
                             & 2           & 3           & 4           & 5           & 6           \\ \hline\hline
2                            & 0.5         & 0.33        & 0.5         & 0.2         & 0.33        \\ \hline
3                            & 1.          & 1.          & 0.5         & 0.8         & 0.67        \\ \hline
4                            & 1.          & 1.          & 0.75        & 1.          & 1.          \\ \hline
\end{tabular}\\
Fraction of targets satisfied for a given link budget and number of targets. 

\label{tab:res}
\vspace{-10pt}
\end{table}

\subsection{Workspace Coverage Analysis}

\begin{figure}[t!]
\centering
    \includegraphics[width=\linewidth]{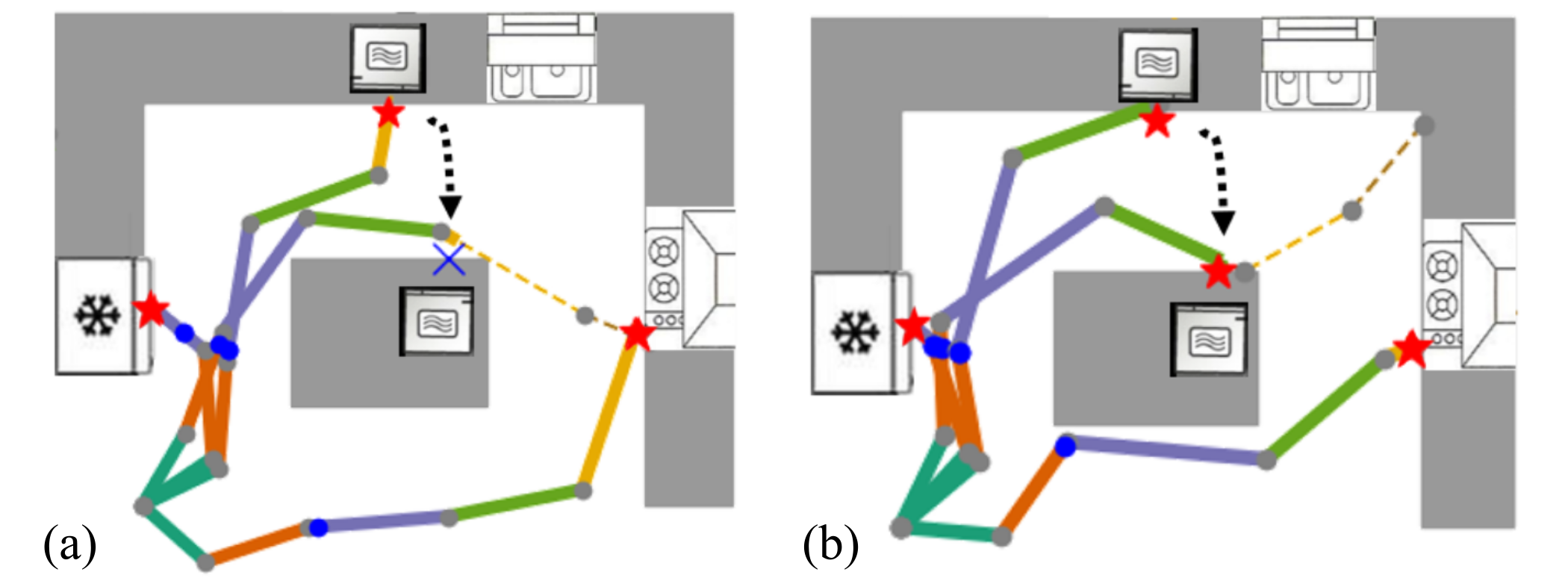}
    \vspace{-15pt}
\caption{Coverage differences between two designs. (a) Without the coverage function (i.e. removing the last term in Equation \ref{eqn:costs}), if the microwave is moved from the counter to the island, it is no longer a reachable target. (b) With the addition of the coverage function, the design has a greater coverage and can still reach the microwave even if it is moved.}
\label{fig:Workspace_Coverage}
\vspace{-5pt}
\end{figure}

We now demonstrate example workspaces for given robot designs. We randomly sampled two thousand points $(t_x,t_y,\phi)$ in $[0, 1] \times [0, 1]\times [-90^{\circ}, 90^{\circ}]$, and computed the ratio of reachable targets to the total number of sampled points in the target space. The coverage computation yields a 33.35\% success rate. The design optimized without the coverage function yields a lower success rate, 27.15\%. Fig. ~\ref{fig:Workspace_Coverage} illustrates the difference between two designs in an environment representing a kitchen, including an island as an obstacle. 


\subsection{Multiple Sequential Targets}

\begin{figure}[t!]
\centering
    \includegraphics[width=\linewidth]{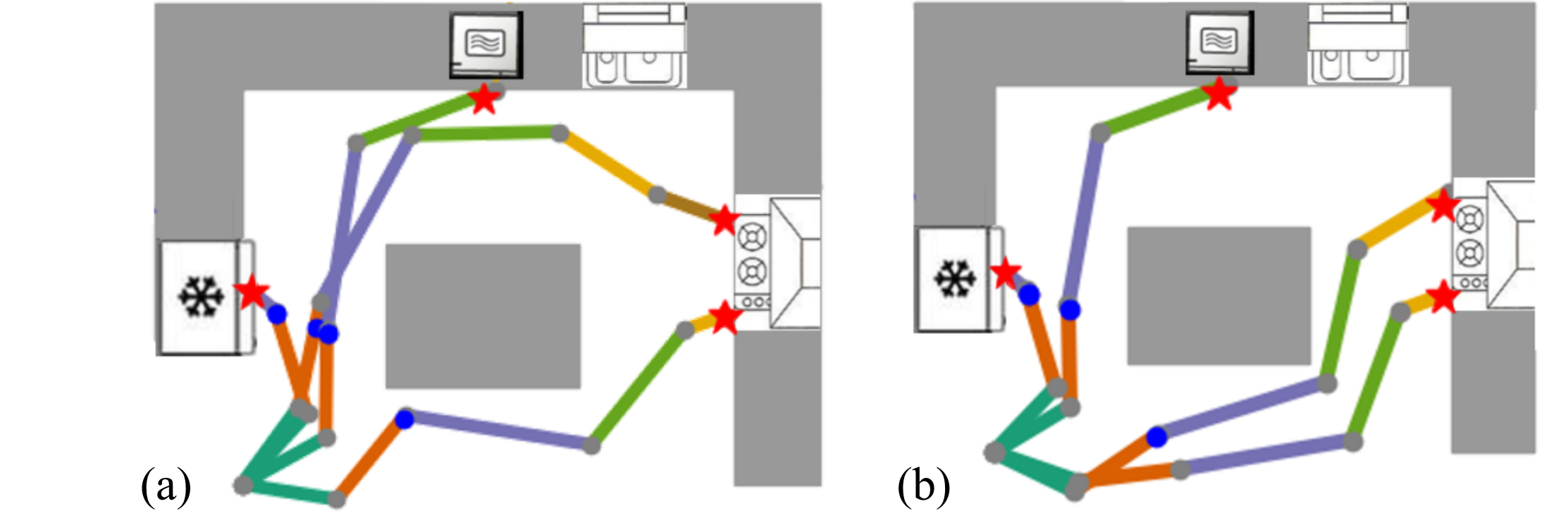}
    \vspace{-15pt}
\caption{Different sequences of reaching the four targets. The point of retraction is marked by the blue dot. (a) Fridge, top of stove, bottom of stove, microwave. (b) Bottom of stove, top of stove, microwave, fridge.}
\label{fig:sequential_targets}
\vspace{-15pt}
\end{figure}

Once the design is optimized and fabricated, the robot can reach a sequence of targets through a series of retracting, growing, and bending without resetting to the base. However, different target orderings may result in different strategies due to obstacles. We implement a simple method that first interpolates the two configurations $\bm{q}_A$ and $\bm{q}_B$ solved independently for two consecutive targets. We check all collision points between obstacles and interpolated configurations and select the closet collision point $\bm{p}_c$ from the home base. We then retract $\bm{q}_A$ until the robot is within $|\bm{p}_c|$-radius from the home base. Once the robot retracts and bends over to configuration $\bm{q}_B$, it can then grow as needed to reach the next target. Fig.~\ref{fig:sequential_targets} shows a 2D visualization of a kitchen space, along with the configurations to reach each target. 


\subsection{Trade-off in Design}
\begin{figure}[t]
\vspace{1.5mm}
\centering
    \includegraphics[width=0.9\linewidth]{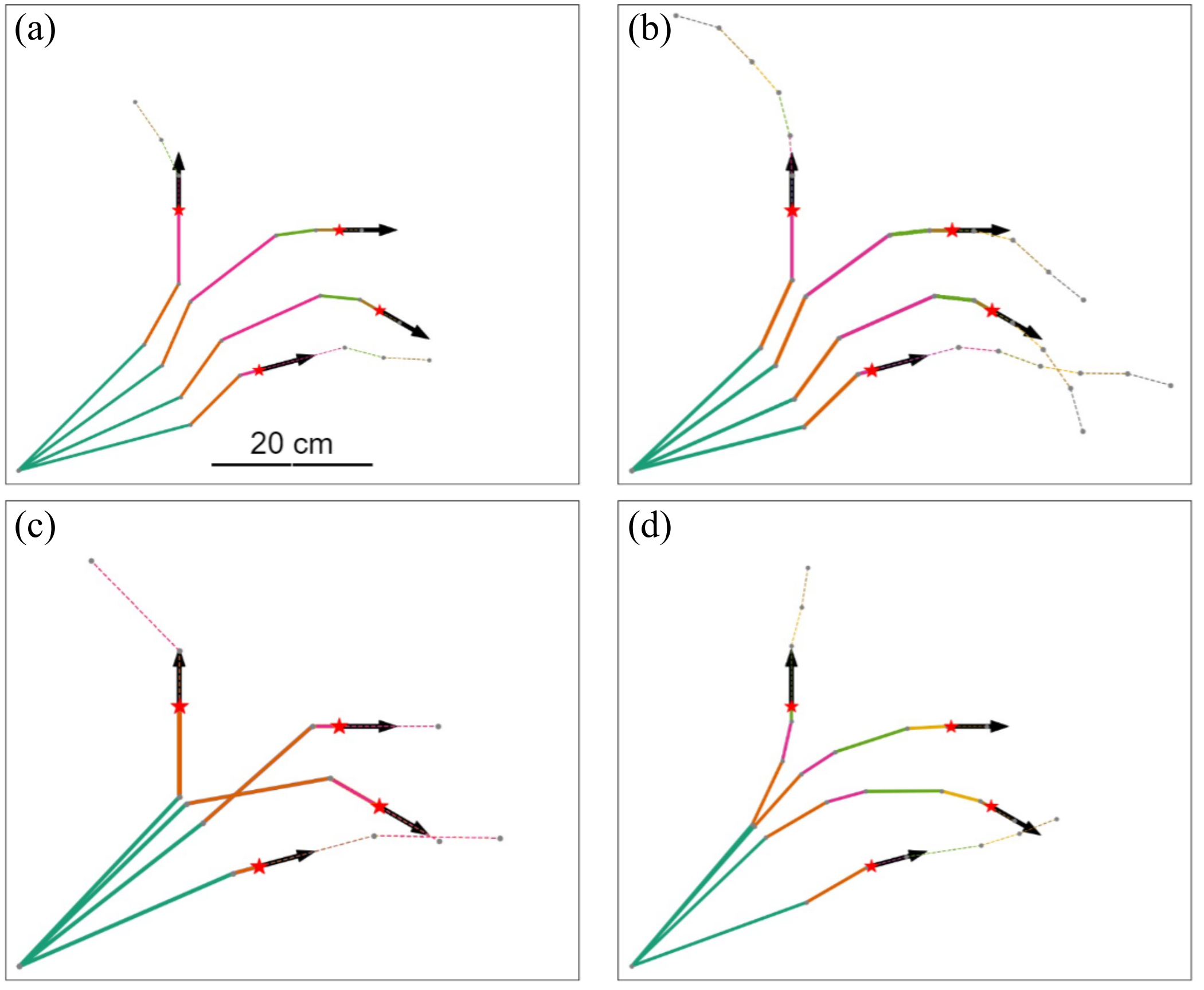}
\caption{Alternative designs for varying hardware constraints and link budgets. (a) The original design, described in Sec.~\ref{Eval}-A.  (b) Using a budget of 8 links yields approximately the same solution: only 5 of them are used. (c) If the robot were able to bend $45^{\circ}$, only 3 links are necessary. (d) If the robot were only able to bend $15^{\circ}$, 6 links are necessary.} \label{fig:des_comp}
\vspace{-10pt}
\end{figure}

To demonstrate the utility of the algorithm as a tool to understand the various designs that could be built given different constraints, we showcase different designs for the same set of targets as those introduced in Section \ref{Eval}-A. The various designs are shown in Fig.~\ref{fig:des_comp}. Specifically, we compare our original design (Fig.~\ref{fig:des_comp}(a)) to the resulting design if the link budget, $n$, is increased from 5 to 8 (Fig.~\ref{fig:des_comp}(b)) and find that the solution is approximately the same, and only 5 of the 8 links are used. In Fig.~\ref{fig:des_comp}(c), a design is shown for a robot that is able to bend $45^{\circ}$ instead of $30^{\circ}$; the solution requires only 3 links. Conversely, a robot being able to bend only $15^{\circ}$ would require 6 links, as shown in Fig.~\ref{fig:des_comp}(d).


\section{Conclusions and Future Work}

In this work, we presented a method to optimize the design of a soft growing vine robot to reach a set of given targets at specified approach angles, while avoiding the obstacles and maximizing the space coverage, such that the design can also be used to achieve other targets not originally specified.  A sample problem of 4 targets generated a design with 5 links, which was fabricated and demonstrated to achieve the task. These results are a key step toward low-cost, bespoke soft robots for user-defined tasks in home environments.

In the future, we plan to achieve the following objectives: extend the optimizer to 3D scenarios; exploit other optimization methods to explore the research space; implement advanced obstacle-avoidance algorithms for navigation in cluttered environments; and improve the fabrication process to enable a faster, tighter loop between robot design and fabrication.





\vspace{-10pt}




\bibliographystyle{IEEEtran}
\bibliography{References,CHARMBib}

\end{document}